\documentclass[letterpaper]{article} 
\usepackage{aaai25}  
\usepackage{times}  
\usepackage{helvet}  
\usepackage{courier}  
\usepackage[hyphens]{url}  
\usepackage{graphicx} 
\usepackage{natbib}  
\usepackage{caption} 
\usepackage{algorithm}
\usepackage{algorithmic}
\usepackage{multirow}
\usepackage{graphicx}
\usepackage{subfigure}
\usepackage{makecell}
\usepackage{xcolor,colortbl}
\usepackage{color}
\usepackage{booktabs}
\usepackage{amsfonts}
\usepackage{amsmath}

\usepackage{newfloat}
\usepackage{listings}
\DeclareCaptionStyle{ruled}{labelfont=normalfont,labelsep=colon,strut=off} 
\floatstyle{ruled}
\newfloat{listing}{tb}{lst}{}
\floatname{listing}{Listing}
\title{Epsilon: Exploring Comprehensive Visual-Semantic Projection for Multi-Label Zero-Shot Learning}
\author{
    Ziming Liu\textsuperscript{\rm 1}, 
    Jingcai Guo\textsuperscript{\rm 1}\footnote{Corresponding author: Jingcai Guo}, 
    Song Guo\textsuperscript{\rm 2}, 
    Xiaocheng Lu\textsuperscript{\rm 2}
}
\affiliations{
    \textsuperscript{\rm 1}Department of Computing, The Hong Kong Polytechnic University, Hong Kong SAR, China\\
    \textsuperscript{\rm 2}Department of CSE, Hong Kong University of Science and Technology, Hong Kong SAR, China\\
    ziming.liu@connect.polyu.hk, jc-jingcai.guo@polyu.edu.hk, xiaocheng.lu@connect.ust.hk, songguo@cse.ust.hk
}

\begin{document}

\maketitle

\begin{abstract}
This paper investigates a challenging problem of zero-shot learning in the multi-label scenario (MLZSL), wherein the model is trained to recognize multiple unseen classes within a sample (e.g., an image) based on seen classes and auxiliary knowledge, e.g., semantic information. 
Existing methods usually resort to analyzing the relationship of various seen classes residing in a sample from the dimension of spatial or semantic characteristics and transferring the learned model to unseen ones. 
However, they neglect the integrity of local and global features. Although the use of the attention structure will accurately locate local features, especially objects, it will significantly lose its integrity, and the relationship between classes will also be affected. Rough processing of global features will also directly affect comprehensiveness. This neglect will make the model lose its grasp of the main components of the image. Relying only on the local existence of seen classes during the inference stage introduces unavoidable bias. 
In this paper, we propose a novel and comprehensive visual-semantic framework for MLZSL, dubbed \textit{Epsilon}, to fully make use of such properties and enable a more accurate and robust visual-semantic projection. In terms of spatial information, we achieve effective refinement by group aggregating image features into several semantic prompts. It can aggregate semantic information rather than class information, preserving the correlation between semantics. In terms of global semantics, we use global forward propagation to collect as much information as possible to ensure that semantics are not omitted. 
Experiments on large-scale MLZSL benchmark datasets \textit{NUS-Wide} and \textit{Open-Images}-v4 demonstrate that the proposed \textit{Epsilon} outperforms other state-of-the-art methods with large margins.
\end{abstract}

%

\section{Introduction}

With the continuous advances in science and technology, the ability and ways human beings obtain information have greatly improved. 
In recent years, deep learning classification networks~\cite{simonyan2014very, szegedy2015going} driven by computational intelligence have sprung up and made rapid progress. 
Image classification datasets designed for single labels in the past are less challenging for deep learning models. Since an arbitrary image may contain multiple objects, the task of multi-label image classification~\cite{gong2013deep, yu2014large, weston2011wsabie} has come into view because the task is challenging, and it is closer to the real life of a human. 
With the continuous improvement of deep neural networks, multi-label image classification models based on deep learning have emerged~\cite{wang2016cnn, durand2019learning, gao2021learning, feng2019collaboration}. Their research focuses on how to effectively process the context information in the image, or build a graph model or other topological models to handle complex dependencies between labels~\cite{chen2019multi}. In recent years, researchers have realized that the attention mechanism has been introduced to allow the model to focus more selectively on information-rich regions and has achieved remarkable results~\cite{cheng2021mltr}. 
\begin{figure}[t]
    \centering
    \includegraphics[width=0.44\textwidth]{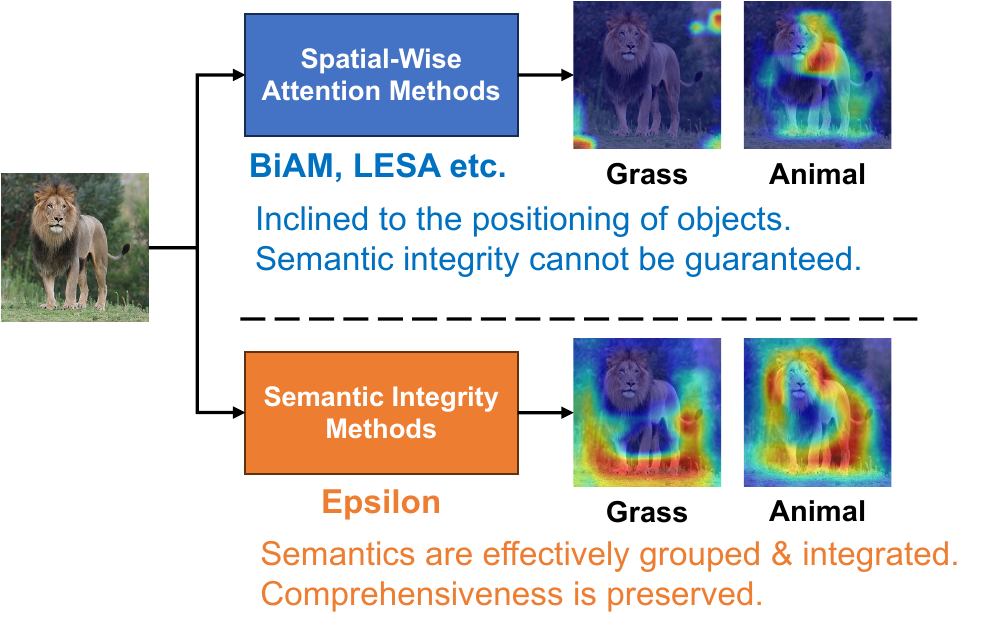}
    \caption{Comparison of attention between our proposed \textit{Epsilon} and traditional spatial attention-based models. We can see that the proposed model is stronger than the contrasting \textit{BiAM} and \textit{LESA} in terms of completeness of class semantics. (zoom in for a better view) }
    \label{fig1}
    \vspace{-5px}
\end{figure}

However, the above multi-label image classification methods all have a common problem, that is, these models cannot perceive categories that were not present during training. Therefore, zero-shot learning (ZSL) has become a solution to this problem~\cite{akata2016multi, deutsch2017zero, frome2013devise}. 
ZSL is essentially a multi-modal transfer learning, that is, an algorithm that predicts unseen classes during testing through the learning of seen classes. Therefore, it is different from the tasks and challenges faced by traditional supervised learning. In fact, existing ZSL models have achieved substantial success in single object prediction~\cite{kodirov2017semantic, xian2017zero, guo2020novel}. 
Compared with traditional ZSL, generalized ZSL is closer to the actual application scenario. Generalized ZSL means that not only unseen classes need to be predicted, but also all seen classes should be detected during testing. However, in practical problems, there are often multiple seen objects and unseen objects in a picture. At this time, the traditional ZSL model will cause serious performance loss and even cannot work effectively. Therefore, multi-label zero-shot learning (MLZSL) has entered the research field as a more complex and practical research direction. 

There have been some works focusing on MLZSL tasks and obtained some achievements with decisive progress in recent years~\cite{norouzi2014zero, mensink2014costa, zhang2016fast}. 
Among them, Zhang~\textit{et al.}~\cite{zhang2016fast} and Ben~\textit{et al.}~\cite{ben2021semantic} calculate the principal direction of its feature vectors for each test image and then sorts the similarity of labels that exist near that principal direction. \textit{LESA}~\cite{huynh2020shared} and \textit{BiAM}~\cite{narayan2021discriminative}, on the other hand, performs multi-label prediction tasks from another perspective, which strengthens the feature response of each class through attention mechanisms. ~\cite{liu20232} innovatively handles the semantic relationship between seen and unseen classes from the perspective of channel response, but this approach is not precise enough for the processing of spatial features.

In response to the problems encountered in the MLZSL method mentioned above, the first problem is that previous methods did not effectively utilize the relationship between local and global features of images. For example, \textit{SDL}~\cite{ben2021semantic} and \textit{Fast0Tag}~\cite{zhang2016fast} only focus on the principal vector of the image, ignoring the supplement of local features to global information. However, \textit{BiAM}~\cite{narayan2021discriminative} simply concatenates global and local features and then performs the classification part. \textit{LESA}~\cite{huynh2020shared} uses the attention sharing of local features to replace the global features. In the multi-label environment, global features are the most important for determining the main components in an image. This ensures that the semantic integrity is not destroyed. As shown in Figure~\ref{fig1}, comparing the attention of the two classes of `grass' and `animal', it is not difficult to see that the overall semantics of the spatial attention-based method are not comprehensive. The consequence is that it will affect the model's learning of the seen class and its ability to transfer to the unseen class. Similar problems occur extremely frequently, especially for semantic extraction of scene classes. Therefore, we urgently need to locally group and integrate the semantics within the image and expand the diversity of global features to make the semantic information corresponding to the image more comprehensive, thereby helping the model to better perform transfer learning. Based on this, we proposed \underline{\textbf{E}}xploring Com\underline{\textbf{p}}rehen\underline{\textbf{si}}ve Visua\underline{\textbf{l}}-Semantic Projecti\underline{\textbf{on}} for Multi-Label Zero-Shot Learning (\textit{\textbf{Epsilon}}) to try our best to find semantic descriptions corresponding to all classes in the picture.

The main contributions of this paper can be summarized as the following three-fold:
\begin{enumerate}
    \item We propose to use visual prompt learning to group and aggregate features, which solves the loss of details caused by traditional methods using spatial attention and greatly ensures semantic integrity.
    \item We use the Global Forward Propagation (GFP) module to greatly enrich the diversity of global features and improve the global information richness. 
    \item Both modules work together on the fused features. Extensive experiments on \textit{NUS-Wide} and \textit{Open-Images}-V4 datasets demonstrate the effectiveness of our method against other state-of-the-art MLZSL models.
\end{enumerate}

\begin{figure*}
    \centering
    \includegraphics[width=0.90\textwidth]{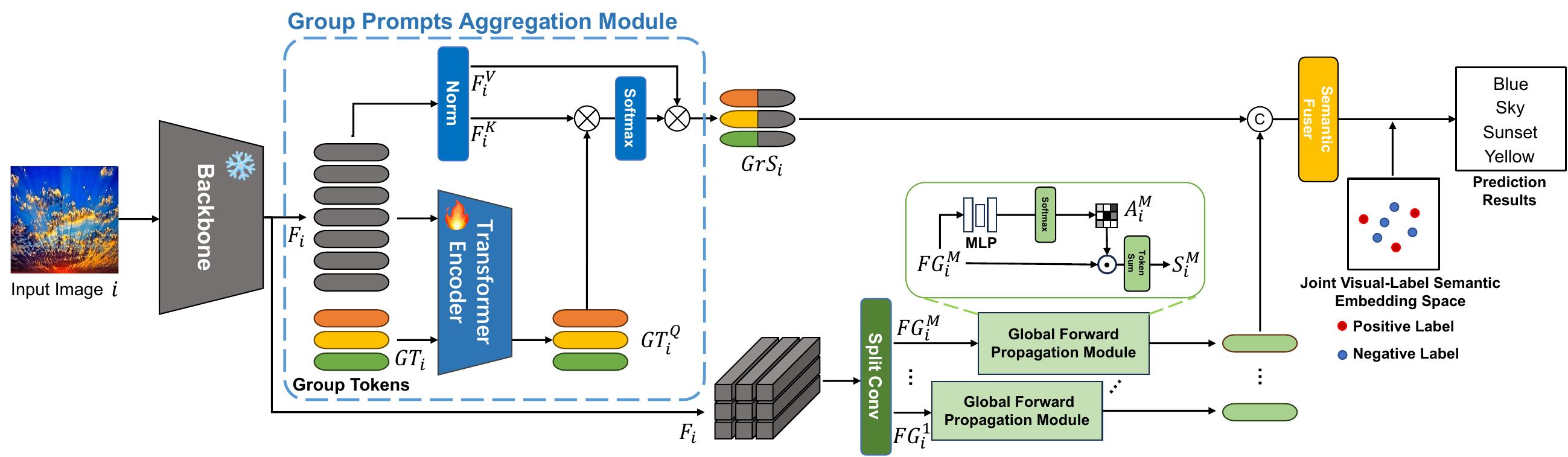}
    \caption{Pipeline of \textit{Epsilon}. The feature representation of the image is first obtained through a pre-trained backbone network that is frozen and cannot be updated. The image features are then applied to the Group Prompts Aggregation Module (GPA Module), which represents local semantic generation, and the Global Forward Propagation Module (GFP Module), which represents global diversity semantic generation. Finally, the output of the two modules is integrated to obtain the complete semantics. (zoom in for a better view)}
    \label{fig2}
    \vspace{-10px}
\end{figure*}

\section{Related Work}
MLZSL presents a more complex challenge than single-label learning due to the unpredictability of the number of labels assigned to each image, thus demanding the model's capacity to weigh multiple unseen labels simultaneously. 
Several methods have emerged to address the challenge. 
Norouzi~\textit{et al.} \cite{norouzi2014zero} developed a model for multi-label zero-shot learning by explicitly partitioning the image and semantic embedding spaces and using a convex combination of label embedding vectors to map the image. Zhang~\textit{et al.} \cite{zhang2016fast} proposed a more general and fast model based on word vectors of ranked relevant labels. 
Lee~\textit{et al.} \cite{lee2018multi} posited that knowledge graphs offer a framework for linking different labels in multi-label environments. He is also the first model to introduce graph structure into multi-label zero-shot learning. 
Attention-based methods such as LESA \cite{huynh2020shared} and Narayan~\textit{et al.} \cite{narayan2021discriminative} have shown promising results. Their approaches leverage the attention-sharing mechanism and bi-layer attention module to focus on key areas of different labels and global context information, respectively. Compared with past methods, they pay more attention to the mining of local semantics and the sharing of features between classes. 
Although Ben~\textit{et al.} \cite{ben2021semantic} utilized the diversity of semantics and embedding matrices to enhance the ability of multi-label zero-shot learning, such methods require analysis in real-world scenarios. Nevertheless, these approaches fall short of exploring the feature channels' response to various classes and only stay at the two-dimensional ($H \times W$) level. 
Liu~\textit{et al.} \cite{liu20232}, starting from the perspective that excessive mining of spatial features will cause loss to image semantics, creatively proposed to find inter-class relationships from the perspective of channel response. Channel responses can also be encoded into effective semantic information in images, which greatly improves model efficiency. 
Gupta~\textit{et al.}~\cite{gupta2023generative} creatively extended the GAN model of single-label zero-shot learning to a multi-label environment. And the multi-label information is fused at the attribute-level and feature-level respectively, and finally summarized at the cross-level.

\section{Methods}
\subsection{Preliminary Problem Setting}
Firstly, it is necessary to clearly define the purpose of the task of MLZSL. Let the batch size be $n$, for the images input into the model $\left \{ \left ( I_1, Y_1\right ),\dots, \left ( I_i, Y_i\right ), \dots, \left ( I_n, Y_n\right )\right \} $, where $I_i$ is the $i$-th image in the input training set, and the corresponding $Y_i$ represents the label of the input image $i$. These labels in the training set are also called `seen labels'. Like ZSL, MLZSL does not overlap the labels of the training and testing sets in terms of label distribution. Let's define the set of all labels in the dataset as $\mathcal{C}$, the set of seen labels as $\mathcal{C}_s$, and the set of unseen labels as $\mathcal{C}_u$. The distribution relationship of labels in the dataset can be described as $\mathcal{C} = \mathcal{C}_s\cup\mathcal{C}_u$. $\mathcal{C}_s$ is mainly used for training sets, while $\mathcal{C}_u$ mainly appears during the testing process. During the MLZSL testing process, for the input image $I_u$, the output prediction result $y_u$ should be $y_u \subset \mathcal{C}_u$. In the generalized MLZSL task, the output prediction result $y_u$ must include `seen labels', which is $y_u \subset \mathcal{C}$.

\subsection{Group Prompts Aggregation Module}
In order to obtain finer-grained spatial features, an effective semantic grouping of spatial features is performed. In this section, we introduce the concept of feature grouping. The specific structure of the model is shown in Figure~\ref{fig2}. First, for the input image $I_i$, we obtain the corresponding feature $F_i$ of the image through a pre-trained backbone network (in this paper, we use the pre-trained ViT-B/16 model~\cite{dosovitskiy2020image} as an example.). Subsequently, we designed several group prompts that can be updated, with the number of group prompts being $M$. Here, $M$ is one of the hyper-parameters of the model, and we will discuss it in the experimental session. In order to better group and aggregate image features, we send the image features $F_i\in \mathbb{R}^{B\times N \times D}$ and $GT_i\in \mathbb{R}^{B\times M \times D}$ representing group prompts to the encoder of Transformer model for aggregation, where $N$ is the number of image tokens, $D$ is the feature dimension of the token. It should be noted here that the Transformer's Encoder can be updated. This is because we need the model itself to learn how to integrate features rather than just relying on updating the prompts themselves. 
The above processes are expressed as:
\begin{align}
    Input_i = [GT_i; F_i],
\end{align}
\begin{align}
    GT_i^Q = Encoder(Input_i)[:M, :],
\end{align}
Among them, $[:M, :]$ represents the first M prompts output, and $Encoder$ represents the encoder of the Transformer model. $GT_i^Q\in \mathbb{R}^{B\times M \times D}$ is the aggregated image feature we obtained. However, the above operations can only roughly group features. 

In order to obtain more refined group features for subsequent encoding of semantic information, we recombine these updateable group prompts $GT_i^Q$ with the original features $F_i$. Here, $GT_i^Q$ serves as query and $F_i$ serves as key and value. This step further refines the semantic information of each group in order to complete visual-semantic projection locally.

\begin{align}
    GrS_i = softmax(\underbrace{ F_i^K \cdot GT_i^Q}_{\mathcal{A}_c}) \cdot F_i^V
\end{align}

Among them, $GrS_i\in \mathbb{R}^{B\times M \times D}$ is the group semantic vector, which is a clustering of local semantic information.

\subsection{Global Forward Propagation Module}
In existing methods, the extraction of global features is rough. Most methods use the top-level features and design some channel (or global) attention structures to achieve this. The above approach ignores the diversity of global information. In fact, the global information itself should have sufficient information abundance to reflect its comprehensiveness. Therefore, we designed the Global Forward Propagation Module (GFP Module) to implement an extraction strategy for richer global information. This new structural design feeds existing global features into multiple modules respectively and uses the personalized global weights obtained by the modules to enhance global information from multiple angles. The specific structure is shown in Figure~\ref{fig2}. We first encode the feature $F\in \mathbb{R}^{B \times N \times D}$ into $M$ feature blocks, respectively $FG_i^1 , FG_i^2, \cdots, FG_i^M \in \mathbb{R}^{B \times N \times D}$. Then, $M$ identical modules are applied to each feature block. The specific implementation process of the module is as follows:
Taking $FG_i^M\in \mathbb{R}^{B \times N \times D}$ as an example, we first use a multi-layer perceptron (MLP) to re-represent the features. Then, the features are fed into softmax to obtain the weight representation of each feature point in the feature map $A_i^M\in \mathbb{R}^{B \times N \times D}$. The reason for this operation is that we need the GFP module to improve the diversity of global features during the forward propagation process. At the same time, we must also ensure that important global information is given the necessary prominence. Therefore, this personalized weight learning and structure design meets our needs. After obtaining $A_i^M$, we perform a dot product with the original input feature $FG_i^M$. Then, all features are summed to obtain the global semantics $S_i^M\in \mathbb{R}^{B \times D}$ corresponding to $FG_i^M$. The above processes are expressed as:
\begin{align}
    W_i^M = MLP(FG_i^M),
\end{align}
\begin{align}
    A_i^M = softmax(W_i^M),
\end{align}
\begin{align}
    S_i^M = Sum(FG_i^M \cdot A_i^M).
\end{align}
Among them, $\cdot$ means the dot product operation. After obtaining all the global semantics $GoS_i = {S_i^1, \dots, S_i^M}$, we use the concatenate operation to combine $GoS_i$ and $GrS_i$ and get $GS_i\in \mathbb{R}^{B \times M \times 2D}$. This idea is mainly referred to Ben~\textit{et al.} \cite{ben2021semantic} and Zhang~\textit{et al.} \cite{zhang2016fast}, which both use the principal vector of the image as the semantic information. Although our GFP module has some structural similarities with the traditional channel attention model, our distribution of weights is not limited to the channel dimension. In addition, since traditional attention models tend to let the model learn a global weight distribution, this will cause the global features of different images to lose diversity. Our multi-head approach enriches this diversity. 

After concatenating the semantic information, we pass a Semantic Fuser, which is a linear layer that helps change the length of the existing semantics so that it is the same as the length of the class semantics in the semantic embedding space, which facilitates the subsequent prediction process. The process is expressed as: 
\begin{align}
    GS_i = Concat(GoS_i, GrS_i),
\end{align}
\begin{align}
    S_i = Linear(GS_i),
\end{align}

\subsection{Loss Function}
Our loss function is inspired by ~\cite{zhang2016fast,ben2021semantic}. For each input image $I_i$, the semantic group of the corresponding image is $\mathcal{S}_i \in \mathbb{R}^{M\times d_w}$, where $d_w$ is the length of label embeddings. This semantic vector group is used to predict unseen classes during the testing process. Therefore, the basic calculation process of the loss function includes the judgment of class similarity. It is necessary to try to improve the ranking of the classes that appear in the input image and reduce the similarity ranking of the classes that do not appear, 
\begin{equation}
    \tau_{jk}=\max\left(n_j\mathcal{S}_i\right) - \max\left(p_k\mathcal{S}_i\right),
\end{equation}
where $n_j$ is the word vector of the seen classes that do not appear, and $p_k$ is the word vector of the seen classes that appear in the input image. In order to maximize the distance between classes that appear and those that do not, we use the maximum value for calculation. At the same time, this approach can ensure sufficient semantic diversity. Therefore, one of the main components of the loss function, namely the ranknet loss~\cite{zhang2016fast}, is shown in the following function:
\begin{equation}
    \mathcal{L}_{r} = \alpha\sum_j\sum_k\log \left(1+e^{\tau_{jk}}\right),
\end{equation}
where $\alpha=\left(\left|T\right|\left|\bar{T}\right|\right)^{-1}$. $\left|T\right|$ and $\left|\bar{T}\right|$ denote the number of seen classes that appear or not, respectively. The hyper-parameter $\alpha$ is used to normalize the ranknet loss. 

In addition, in order to reduce the difficulty of the model in predicting hard-positive samples with a high diversity of labels, we suggest introducing a new weight like~\cite{ben2021semantic}. Due to the uniqueness of multi-label tasks, each image contains a different number of labels. When a large number of classes appear in the image, it is necessary to make the model sensitive to label diversity and have a tendency towards hard-positive tasks during the training process.
\begin{equation}
    \omega = 1 + \sum_i var(Y_i). 
\end{equation}
In addition, the loss function of the model introduces the second main component, namely the regularized loss function. Its purpose is to construct correlations between input semantic vectors.
\begin{equation}
    \mathcal{L}_{reg} = \left \| \sum_M var(\mathcal{S}_i^M) \right \|_1.
\end{equation}
Finally, our complete loss function $\mathcal{L}$ is as follows:
\begin{equation}
    \mathcal{L} = \frac{1}{B}\sum_{i=1}^B\left(w \cdot (1-\lambda)\mathcal{L}_{r}(\mathcal{S}_i, Y_i) + \lambda \mathcal{L}_{reg}(\mathcal{S}_i)\right),
\end{equation}
where $\lambda$ is the weight of the regularized loss function, and $B$ is the batch size.

\begin{table*}[htbp]
  \centering
  \caption{Comparison of different models for multi-label ZSL and GZSL tasks on the \textbf{NUS-Wide} dataset. The ``P'' refers to the top-K ``Precision'' and ``R'' refers to the top-K ``Recall''. The best results are shown in {\color{red}\textbf{red}}, and the second are shown in {\color{blue}\textbf{blue}}. }
\resizebox{\textwidth}{!}{
\begin{tabular}{l c ccc c ccc c c ccc c ccc}
    \toprule
     & \multicolumn{8}{c}{\textbf{ZSL}} && \multicolumn{8}{c}{\textbf{GZSL}} \\\cline{2-9}\cline{11-18}
     \textbf{Method} & \multirow{2}{*}{\textbf{mAP}} & \multicolumn{3}{c}{K=3} && \multicolumn{3}{c}{K=5} && \multirow{2}{*}{\textbf{mAP}} & \multicolumn{3}{c}{K=3} && \multicolumn{3}{c}{K=5} \\ \cline{3-5}\cline{7-9}\cline{12-14}\cline{16-18}
     &&\textbf{P} & \textbf{R} & \textbf{F1} && \textbf{P} & \textbf{R} & \textbf{F1} && & \textbf{P} & \textbf{R} & \textbf{F1} && \textbf{P} & \textbf{R} & \textbf{F1} \\\hline
     CONSE~\cite{norouzi2014zero} & 9.4 & 17.5 & 28.0 & 21.6 && 13.9 & 37.0 & 20.2 && 2.1 & 11.5 & 5.1 & 7.0 && 9.6 & 7.1 & 8.1 \\
     LabelEM~\cite{akata2016label} & 7.1 & 15.6 & 25.0 & 19.2 && 13.4 & 35.7 & 19.5 && 2.2 & 15.5 & 6.8 & 9.5 && 13.4 & 9.8 & 11.3 \\
     Fast0Tag~\cite{zhang2016fast} & 15.1 & 22.6 & 36.2 & 27.8 && 18.2 & 48.4 & 26.4 && 3.7 & 18.8 & 8.3 & 11.5 && 15.9 & 11.7 & 13.5 \\
     Kim~\textit{et al.}~\cite{kim2018bilinear} & 10.4 & 20.9 & 33.5 & 25.8 && 16.2 & 43.2 & 23.6 && 3.7 & 17.9 & 7.9 & 10.9 && 15.6 & 11.5 & 13.2 \\
     LESA ApC~\cite{huynh2020shared} & 12.9 & 20.0 & 31.9 & 24.6 && 15.7 & 41.9 & 22.9 && 2.6 & 10.4 & 4.6 & 6.4 && 9.1 & 6.7 & 7.7 \\
     LESA~\cite{huynh2020shared} & 19.4 & 25.7 & 41.1 & 31.6 && 19.7 & 52.5 & 28.7 && 5.6 & 23.6 & 10.4 & 14.4 && 19.8 & 14.6 & 16.8 \\
     BiAM~\cite{narayan2021discriminative} & 25.8 & 26.0 & 41.6 & 32.0 && {\color{blue}\textbf{20.2}} & {\color{red}\textbf{53.9}} & 29.4 && 8.9 & 25.2 & 11.3 & 15.5 && 21.8 & 16.0 & 18.5 \\
     SDL~\cite{ben2021semantic} & {\color{blue}\textbf{25.9}} & 24.2 & 41.3 & 30.5 && 18.8 & 53.4 & 27.8 && {\color{blue}\textbf{12.1}} & 27.7 & {\color{red}\textbf{13.9}} & 18.5 && 23.0 & {\color{red}\textbf{19.3}} & 21.0 \\
     Gen-MLZSL~\cite{gupta2023generative} & 25.7 & {\color{blue}\textbf{26.6}} & {\color{red}\textbf{42.8}} & {\color{blue}\textbf{32.8}} && 20.1 & {\color{blue}\textbf{53.6}} & 29.3 && 8.9 & {\color{blue}\textbf{30.9}} & {\color{blue}\textbf{13.6}} & {\color{blue}\textbf{18.9}} && {\color{blue}\textbf{26.0}} & {\color{blue}\textbf{19.1}} & {\color{blue}\textbf{22.0}} \\
     (ML)$^2$P-Encoder~\cite{liu20232} & 29.4 & - & - & {\color{blue}\textbf{32.8}} && - & - & {\color{blue}\textbf{32.3}} && 10.2 & - & - & 16.8 && - & - & 19.2 \\
     Ours & {\color{red}\textbf{33.2}} & {\color{red}\textbf{31.1}} & {\color{blue}\textbf{41.7}} & {\color{red}\textbf{35.6}} && {\color{red}\textbf{24.9}} & 52.7 & {\color{red}\textbf{33.8}} && {\color{red}\textbf{14.7}} & {\color{red}\textbf{32.8}} & 13.1 & {\color{red}\textbf{18.7}} && {\color{red}\textbf{28.1}} & 18.8 & {\color{red}\textbf{22.5}} \\
     \toprule
\end{tabular}
}
\label{tab1}%
\vspace{-10px}
\end{table*}

\section{Experiments}
\subsection{Experimental Setup}
\noindent\textbf{Datasets:} The \textit{NUS-Wide} dataset~\cite{chua2009nus} contains approximately 270,000 images and a total of 1,006 labels. Among them, 81 labels manually annotated by humans will serve as labels for `unseen classes'. At the same time, these labels will also serve as `ground-truth' labels in the multi-label classification task. The remaining 925 labels were automatically extracted from Flickr users' manual annotations of these images, where they will be used as labels for `seen classes'. This setting is similar with~\cite{huynh2020shared, ben2021semantic}. 
\noindent Another dataset is called the \textit{Open-Images}-V4 dataset, which is much larger than the \textit{NUS-Wide} dataset. This dataset contains approximately 9.2 million images, of which approximately 9 million are used as the training set. The training set contains a total of 7,186 labels, ensuring that each label appears at least 100 times in the training set. These will be considered as labels for `seen classes'. In addition, the dataset also contains 125,456 test images and 400 `unseen classes' labels. These labels are derived from the other 400 most frequent labels that did not appear in the training set, which appeared at least 75 times. The setting of \textit{Open-Images}-V4 dataset is similar with~\cite{huynh2020shared, narayan2021discriminative}.

\noindent\textbf{Evaluation Metrics:} In order to ensure the unbiased comparison and scientific evaluation of the metrics themselves using our proposed method and comparison method, we will use the two most commonly used evaluation metrics in MLZSL and Multi-label Classification tasks, namely the mean Average Precision (mAP) and F1-Score~\cite{veit2017learning, huynh2020shared}. 
Among them, F1-Score is the harmonic mean of the precision and recall. \textit{top-K} F1-Score is an evaluation metric used in classification tasks to measure the accuracy of the model in predicting labels. 
mAP is a class-wise evaluation indicator that is used to reflect the accuracy of image image retrieval from unseen labels. 

\noindent\textbf{Implementation Details:} As for the selection of backbone network, we choose the pre-trained ViT-B/16~\cite{dosovitskiy2020image} as our backbone network. Unlike the comparison method that uses multi-stage training, our method is completely end-to-end training, saving a lot of training time and making the model more concise. We choose the Adam optimizer~\cite{kingma2014adam}as the model's optimizer, which is suitable for large-scale datasets and requires less memory. The weight decay of the Adam optimizer is set to $4e^{-3}$. For the experiments of all the models in the \textit{NUS-Wide} dataset, the entire training process requires a total of 7 epochs with a batch size of 96, and the initial learning rate is set to $1e^{-5}$, and then decreases by $\frac{1}{2}$ at the 4-th epoch. 
In the experiments of the \textit{Open-Images}-V4 dataset, the number of epochs in the training process is set to 7. This is consistent with the settings for other comparison methods. Our optimizer's decay rate, model's learning rate, and batch size remain the same.

\noindent\textbf{Baselines:} 
In terms of baseline methods, we need to compare our proposed new model with all state-of-the-art MLZSL models based on deep neural networks in recent years. The baseline method covers all research directions, including but not limited to attention mechanism, principal direction of feature vectors, etc.
These comparison methods include:
\textit{CONSE}~\cite{norouzi2014zero}, 
\textit{LabelEM}~\cite{akata2016label}, 
\textit{Fast0Tag}~\cite{zhang2016fast}, 
\textit{Kim~\textit{et al.}}~\cite{kim2018bilinear}, 
\textit{LESA Attention per Cluster}~\cite{huynh2020shared}, 
\textit{LESA}~\cite{huynh2020shared}, 
\textit{BiAM}~\cite{narayan2021discriminative},
\textit{SDL}~\cite{ben2021semantic},
Gen-MLZSL~\cite{gupta2023generative},
and \textit{(ML)$^2$P-Encoder}~\cite{liu20232}. 

\subsection{MLZSL performance for \textit{Epsilon}}
We show the performance of both ZSL and GZSL tasks in Table~\ref{tab1}. 
Firstly, for \textit{LESA}~\cite{huynh2020shared} and \textit{BiAM}~\cite{narayan2021discriminative}, as two models with the help of spatial attention mechanism, they are inspired by the multi-label classification methods, and they pay attention to the attention sharing between labels and the personalized attention mechanism respectively, have made enough progress. \textit{SDL}~\cite{ben2021semantic} has a certain hard split on the principal vector compared with \textit{Fast0Tag}~\cite{zhang2016fast}, which objectively improves the multi-label performance. However, \textit{(ML)$^2$P-Encoder}~\cite{liu20232} starts from the perspective of the channel and uses the response for semantic encoding, achieving sub-optimal results in the two main indicators of F1-Score and mAP. Judging from the results of the GZSL task, \textit{SDL}~\cite{ben2021semantic} uses image principal vectors to generate more generalized semantics, which has absolute advantages in the GZSL environment. However, due to the design of Group Prompts, our method also learns sufficient semantic information through training. At the same time, more comprehensive global semantics also help us obtain more generalized image information, which greatly improves the ZSL and GZSL performance of the model, whether $K=3$ or $K=5$.

\begin{table*}[htbp]
  \centering
  \caption{Comparison of different models for multi-label ZSL and GZSL tasks on the \textbf{Open-Images}-V4 dataset. The ``P'' refers to the top-K ``Precision'' and ``R'' refers to the top-K ``Recall''. Best results are shown in {\color{red}\textbf{red}}, and the second are shown in {\color{blue}\textbf{blue}}. }
\resizebox{\textwidth}{!}{
\begin{tabular}{l c ccc c ccc c c ccc c ccc}
    \toprule
     & \multicolumn{8}{c}{\textbf{ZSL}} && \multicolumn{8}{c}{\textbf{GZSL}} \\\cline{2-9}\cline{11-18}
     \textbf{Method} & \multirow{2}{*}{\textbf{mAP}} & \multicolumn{3}{c}{K=10} && \multicolumn{3}{c}{K=20} && \multirow{2}{*}{\textbf{mAP}} & \multicolumn{3}{c}{K=10} && \multicolumn{3}{c}{K=20} \\ \cline{3-5}\cline{7-9}\cline{12-14}\cline{16-18}
     &&\textbf{P} & \textbf{R} & \textbf{F1} && \textbf{P} & \textbf{R} & \textbf{F1} && & \textbf{P} & \textbf{R} & \textbf{F1} && \textbf{P} & \textbf{R} & \textbf{F1} \\\hline
     CONSE~\cite{norouzi2014zero} & 40.4 & 0.2 & 7.3 & 0.4 && 0.2 & 11.3 & 0.3 && 43.5 & 2.4 & 2.8 & 2.6 && 1.7 & 3.9 & 2.4 \\
     LabelEM~\cite{akata2016label} & 40.5 & 0.2 & 8.7 & 0.5 && 0.2 & 15.8 & 0.4 && 45.2 & 4.8 & 5.6 & 5.2 && 3.7 & 8.5 & 5.1 \\
     Fast0Tag~\cite{zhang2016fast} & 41.2 & 0.3 & 12.6 & 0.7 && 0.3 & 21.3 & 0.6 && 45.2 & 14.8 & 17.3 & 16.0 && 9.3 & 21.5 & 13.0 \\
     LESA ApC~\cite{huynh2020shared} & 40.7 & 0.6 & 22.9 & 1.2 && 0.4 & 32.4 & 0.9 && 44.9 & 15.7 & 18.3 & 16.9 && 9.6 & 22.3 & 13.5 \\
     LESA~\cite{huynh2020shared} & 41.7 & 0.7 & 25.6 & 1.4 && 0.5 & 37.4 & 1.0 && 45.4 & 16.2 & 18.9 & 17.4 && 10.2 & 23.9 & 14.3 \\
     BiAM~\cite{narayan2021discriminative} & 62.8 & 2.3 & 17.8 & 4.1 && 2.0 & 31.0 & 3.7 && 79.6 & 16.4 & 18.9 & 17.6 && 10.9 & 25.0 & 15.1 \\
     SDL~\cite{ben2021semantic} & 62.9 & {\color{blue}\textbf{6.1}} & {\color{blue}\textbf{47.0}} & {\color{blue}\textbf{10.7}} && {\color{blue}\textbf{4.4}} & {\color{red}\textbf{68.1}} & {\color{blue}\textbf{8.3}} && 75.3 & {\color{blue}\textbf{35.3}} & {\color{blue}\textbf{40.8}} & {\color{blue}\textbf{37.8}} && {\color{blue}\textbf{23.6}} & {\color{blue}\textbf{54.5}} & {\color{blue}\textbf{32.9}} \\
     Gen-MLZSL~\cite{gupta2023generative} & 43.0 & 1.3 & 42.4 & 2.5 && 1.1 & 52.1 & 2.2 && 75.5 & 33.6 & 38.9 & 36.1 && 22.8 & 52.8 & 31.9 \\
     (ML)$^2$P-Encoder~\cite{liu20232} & {\color{blue}\textbf{65.7}} & - & - & 7.5 && - & - & 6.5 && {\color{blue}\textbf{79.9}} & - & - & 27.6 && - & - & 24.1 \\
     Ours & {\color{red}\textbf{68.5}} & {\color{red}\textbf{7.7}} & {\color{red}\textbf{53.2}} & {\color{red}\textbf{13.5}} && {\color{red}\textbf{4.7}} & {\color{blue}\textbf{64.3}} & {\color{red}\textbf{8.8}} && {\color{red}\textbf{81.0}} & {\color{red}\textbf{37.2}} & {\color{red}\textbf{41.8}} & {\color{red}\textbf{39.4}} && {\color{red}\textbf{24.9}} & {\color{red}\textbf{57.2}} & {\color{red}\textbf{34.7}} \\
     \toprule
\end{tabular}
}
  \label{tab2}%
  \vspace{-10px}
\end{table*}
Table~\ref{tab2} shows the performance of ours and other competitive methods on the \textit{Open-Images-V4} test-set. For fairness, we keep the same training epoch as the comparison method. In the ZSL task, due to the limited number of ZSL labels contained in most samples, the precision indicators of all methods are low. However, our model still achieves relatively the best results, both in terms of precision and recall. In the scenario of this data set, it can be seen that the performance of \textit{LESA} and \textit{BiAM} has dropped significantly compared with \textit{SDL}. This is mainly because as the total number of labels increases, the model's demand for generalization performance also increases. This is what traditional multi-label classification models lack and, therefore, perform poorly. The most difficult situation occurs in GZSL, where the number of labels that need to be predicted reaches around 5,000, which is undoubtedly a severe challenge to the model. In this case, our mAP is still the best. In terms of F1-Score performance, we are still 1.6\% and 1.8\% ahead of \textit{SDL}, respectively. 

\begin{figure}[htbp]
    \centering
    \subfigure[$M$] {\includegraphics[width=.4\textwidth]{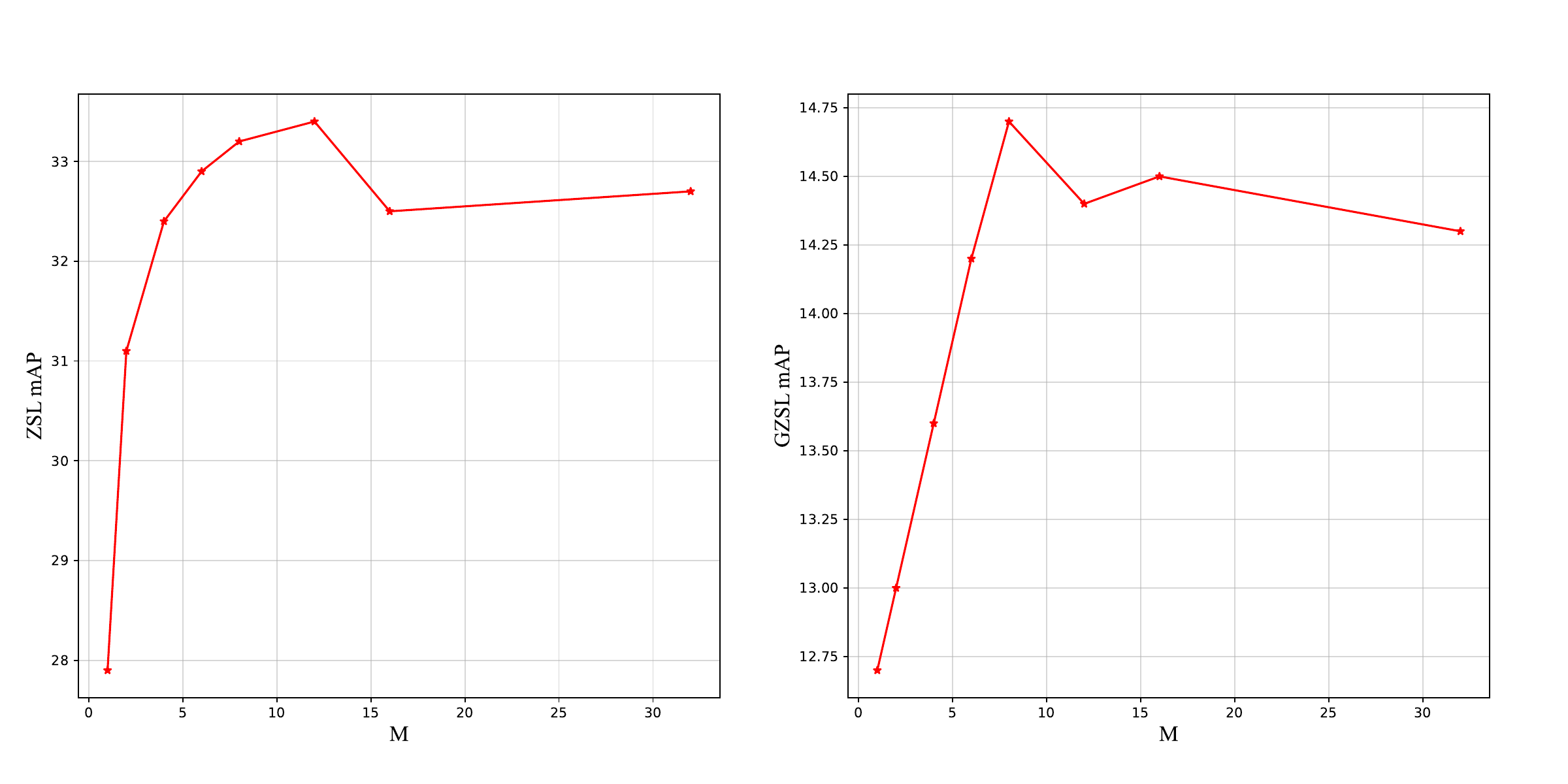}}
    \subfigure[Weights $\lambda$] {\includegraphics[width=.4\textwidth]{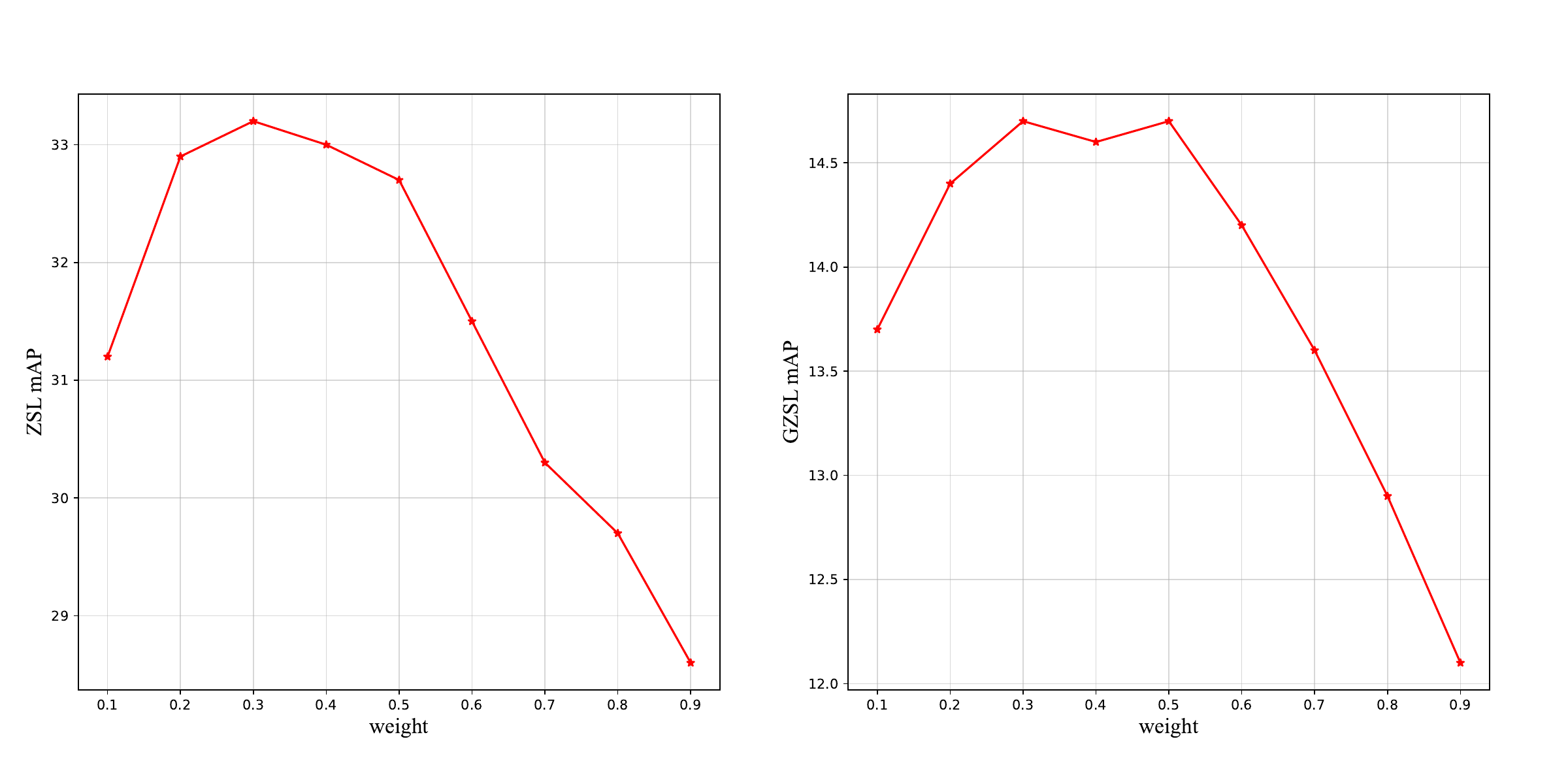}}
    \caption{\textbf{Hyper-parameter selection.} All the experiments are performed on the \textit{NUS-Wide} test-set.}
    \label{fig:Compare}
    \vspace{-10px}
\end{figure}

\begin{table}[htbp]
  \centering
  \caption{\textbf{Ablation study} results. It shows the contribution of the different modules in \textit{Epsilon}. All the models are performed on the test-set of the \textit{NUS-Wide}.}
  \resizebox{.46\textwidth}{!}{
    \begin{tabular}{cr|cccccc|c}
    \multicolumn{2}{c|}{\textbf{Model}} & a     & b  & c  & d &e &f     & ours \\
    \hline
    \multicolumn{2}{r|}{Backbone Only} &   \checkmark    &  \checkmark  &   &   &   &  &  \\
    \multicolumn{2}{r|}{+GPA Module} &   &    & \checkmark       & \checkmark &  &   &\checkmark    \\
    \multicolumn{2}{r|}{+GFP Module} &   &    &   &    &  \checkmark  &  \checkmark       &\checkmark    \\
    \multicolumn{2}{r|}{M=8} &\checkmark       &     &  \checkmark  &   &\checkmark  &  &\checkmark  \\
    \multicolumn{2}{r|}{M=2} &       & \checkmark  &    &  \checkmark        &  &  \checkmark        &   \\
    \hline
    \multirow{2}[1]{*}{\textbf{mAP}} & ZSL    & 28.6 & 27.9  & 32.4 & 30.7 & 31.8 & 30.5 & \bf{33.2}\\
                            & GZSL    & 10.8 & 10.6  & 13.2 & 12.5 & 13.6 & 12.8 &   \bf{14.7}\\
    \end{tabular}}%
  \label{tab3}%
  \vspace{-10px}
\end{table}

\begin{table*}[htbp]
  \centering
  \caption{Models comparison on \textbf{NUS-Wide} dataset under the task of \textbf{Multi-label Classification}. Best results are shown in bold. }
    \scalebox{1}{
    \begin{tabular}{lcccccc|c}
    \toprule
    \textbf{Method}     & \textbf{P(K=3)}  & \textbf{R(K=3)} & \textbf{F1(K=3)} & \textbf{P(K=5)}  & \textbf{R(K=5)} & \textbf{F1(K=5)} &\textbf{mAP} \\
    \midrule
    \textit{Logistic}                     & 46.1   & 57.3   & 51.1  & 34.2  & 70.8  & 46.1  & 21.6 \\
    \textit{WARP}                                & 49.1   & 61.0   & 54.4  & 36.6  & 75.9  & 49.4  & 3.1 \\
    \textit{WSABIE}                          & 48.5   & 60.4   & 53.8  & 36.5  & 75.6  & 49.2  & 3.1 \\
    \textit{Fast0Tag}                           & 48.6   & 60.4   & 53.8  & 36.0  & 74.6  & 48.6  & 22.4 \\
    \textit{CNN-RNN}                             & 49.9   & 61.7   & 55.2  & 37.7  & 78.1  & 50.8  & 28.3 \\
    One Attention per Label          & 51.3   & 63.7   & 56.8  & 38.0  & 78.8  & 51.3  & 32.6 \\
    One Attention per Cluster (M = 10)  & 51.1   & 63.5   & 56.6   & 37.6   & 77.9   & 50.7  & 31.7 \\
    \textit{LESA} (M = 10)                       & 52.3   & 65.1   & 58.0   & 38.6   & 80.0   & 52.0  & 31.5 \\
    \textit{BiAM}                     & -   & -   & 59.6   & -   & -   & 53.4  & 47.8 \\
    \textit{(ML)$^2$P-Encoder}                      & -   & -   & 59.8   & -   & -   & 53.8  & 48.0 \\
    \bf{Ours}                                & \bf{56.3}& \bf{70.0}& \bf{62.4}     & \bf{41.5} & \bf{86.0}& \bf{55.9}     & \bf{61.1} \\
    \bottomrule
    \end{tabular}}%
  \label{tab4}%
  \vspace{-10px}
\end{table*}

\subsection{Hyper-Parameters and Ablation Study}
\noindent \textbf{Hyper-Parameters:} In terms of hyper-parameter selection, our model has two hyper-parameters, $M$ and $\lambda$, as shown in the figure. First of all, we focus on the number of groups $M$. It can be seen that the model we designed does not mean that the larger the $M$, the better the performance. This is because once the features are grouped into too many prompts, the semantic information will be extremely scattered, which will lead to the decline of mAP results. But when the number of $M$ is too small, the semantic diversity cannot be protected. Thus, the semantic output is not comprehensive. Therefore, it can be seen from the experimental results that the model can achieve relatively best results when $M$ is 8. For another hyper-parameter $\lambda$, we also conducted experimental verification. It can be seen that when $\lambda=0.1$, the loss function of the model mostly includes $\mathcal{L}_{r}$, and $\mathcal{L}_{reg}$ has the weakest influence on the loss function. When $\lambda=0.9$, the loss function of the model mostly contains the regular term $\mathcal{L}_{reg}$. From the perspective of the mAP results caused by the change of $\lambda$, adding the regularization term appropriately will help the convergence of the model. From the observation in Figure~\ref{fig:Compare} (b), we can discover that when $\lambda=0.3$, the model can obtain relatively optimal results. 

\noindent \textbf{Ablation Study:} In order to verify the effectiveness of each module of our method and to test the integrity of the method, we designed an ablation study experiment. By comparing models `a' and `b', `c' and `d', and `e' and `f', we can find that the impact of changing hyper-parameter $M$ will not be affected by changes in the model structure. The experimental results of the model when $M=8$ are always stronger than those when $M=2$. In addition, through horizontal comparison, we will find that the addition of each module has a positive effect on the performance improvement of the model's ZSL and GZSL tasks. But they each have their areas of expertise. For example, we can compare the improvements brought by the GPA module and the GFP module to the model between `c' and `e'. We can observe that the GPA module is obviously more beneficial to the ZSL task, while GFP is better at handling the GZSL task. This is because global features generally have stronger generalization capabilities than local group aggregation of features, which is why \textit{SDL}~\cite{ben2021semantic} is slightly better than \textit{BiAM}~\cite{narayan2021discriminative} in GZSL performance. However, local semantics is more effective for learning seen classes, so the ZSL performance of GPA is stronger than the GFP module. From the final results, we clearly know that the best results are produced when the two modules act on the model at the same time.

\subsection{Multi-Label Classification}
As a prediction model, we need to accurately classify seen classes to realize the prediction of unseen classes. Therefore, the multi-label classification ability of the model is also important. Table~\ref{tab4} shows the performance of our model under multi-label learning. For comparison, we added some classic multi-label classification algorithms, including \textit{Logistic Regression}~\cite{tsoumakas2007multi}, \textit{WSABIE}~\cite{weston2011wsabie}, \textit{WARP}~\cite{gong2013deep} and \textit{CNN-RNN}~\cite{wang2016cnn}, and some MLZSL models. We can see from the experimental results that our method has achieved the best results compared with the traditional multi-label learning model in both F1-Score and mAP. Compared with \textit{BiAM}~\cite{narayan2021discriminative} and \textit{(ML)$^2$P-Encoder}~\cite{liu20232}, the current best-performing methods in MLZSL, our model still has considerable advantages. This is because, compared with \textit{BiAM}, our model acquires richer semantic information and has a stronger ability to integrate class features, which can significantly improve the model's classification ability.

\subsection{Qualitative Performance}
\begin{figure}[htbp]
    \centering
    \includegraphics[width=.46\textwidth]{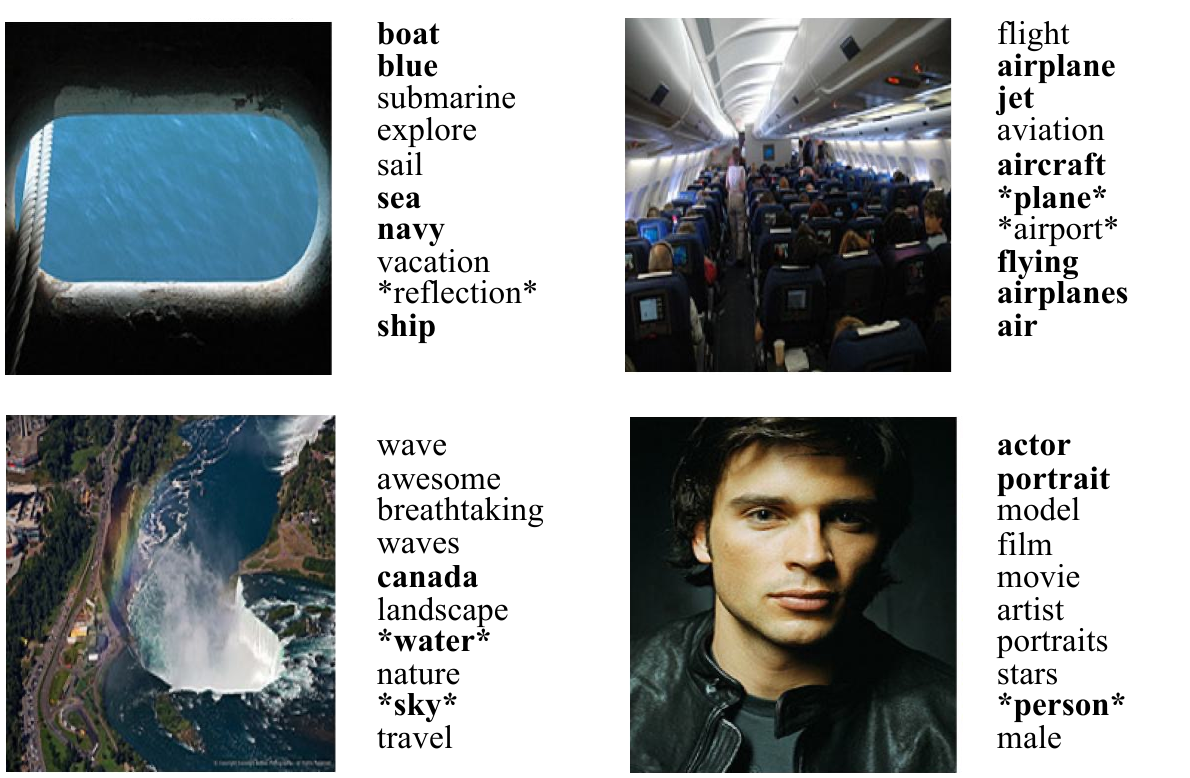}
    \caption{Top-10 labels predicted by \textit{Epsilon} in the case of Generalized MLZSL on NUS-Wide dataset. The asterisk marks indicate unseen labels, while the bold ones indicate successfully predicted seen and unseen labels.}
    \label{fig:quali}
    \vspace{-5px}
\end{figure}

Our model's qualitative performance is shown in Figure~\ref{fig:quali}. For the input image, the label output by the model includes not only seen labels but also unseen labels. Scene semantics is more about the model's perception of global information. In the first picture, humans do not realize that it is a boat sailing on the sea at first glance. However, the model first recognized `boat' and `blue', and then other classes related to its semantics were also easily predicted. In the second picture, judging from the scene information, our model can accurately locate the scene as an aircraft. The scene information in the third picture is more complex, but the model can still recognize `water' and even abstract classes such as `Canada'. The above three pictures clearly show that global semantic diversity gives the model more perspectives to interpret global information. Detailed features rely more on the integration of local semantics. The fourth image accurately identifies the character's gender and identifies him as an actor based on his face and clothing. The success of the above identifications undoubtedly reflects that our model is very effective in both the expression of global semantics and the aggregation of local semantics. 

\section{Conclusion}
This paper designs \textit{Epsilon} to solve the problem of the lack of integrity of local features and rough processing of global features, resulting in the loss of visual-semantic comprehensiveness. In response to the above problems, we first designed the GPA module to improve the integrity of local semantic information by grouping and aggregating spatial features. Then, the GFP module will be used to obtain more diverse and comprehensive global features. Finally, the global and local features are combined as complete semantic information of an image for training. On two publicly available large MLZSL datasets \textit{NUS-Wide} and \textit{Open-Images}-V4, our experimental results far outperform other state-of-the-art models.

\appendix
\section{Supplementary Materials of \textit{Epsilon}}

\begin{table}[htbp]
  \centering
  \caption{Comparison with CLIP-FT for multi-label ZSL task on the \textbf{NUS-Wide} dataset.}
\scalebox{0.83}{\begin{tabular}{lcccc}
    \toprule
    \textbf{Methods} & \textbf{mAP}   & \textbf{F1 (K = 3)}  & \textbf{F1 (K = 5)} \\
    \midrule
     CLIP-FT~\cite{radford2021learning} & 30.5 & 23.5 & 21.7 \\
     Ours & \textbf{33.2} & \textbf{35.6} & \textbf{33.8} \\
     \toprule
\end{tabular}
}
\label{tab1}%
\end{table}

\subsection{Zero-Shot Comparison with CLIP-FT}
In this section, we compare our method with the CLIP~\cite{radford2021learning} model fine-tuned on the NUS-Wide~\cite{chua2009nus} dataset. We use exactly the same backbone network as CLIP. The difference is that CLIP is a classic multi-modal model, and its text branch can provide the model with Open-Vocabulary capabilities. However, in the testing session, it can be seen that our model performs far better than CLIP-FT in the ZSL task. This result fully demonstrates the powerful performance of our model, as well as the ability to learn single-modal knowledge efficiently. However, since CLIP itself has a very powerful text model, the model itself has sufficient additional knowledge. Therefore, as the number of classes increases, its performance will become better.

\begin{table}[htbp]
  \centering
  \caption{Comparison with different backbones in NUS-Wide}
    \scalebox{0.79}{\begin{tabular}{lcccc}
    \toprule
    \textbf{Backbones} & \textbf{Task}  & \textbf{mAP}   & \textbf{F1 (K = 3)}  & \textbf{F1 (K = 5)} \\
    \midrule
    
    \multirow{2}[2]{*}{Ours(ViT-B/16)}& ZSL   & \textbf{33.2}     & \textbf{35.6}    & \textbf{33.8} \\
                                    & GZSL  & \textbf{14.7}     & \textbf{18.7}     & \textbf{22.5} \\
    \cmidrule{2-5}  
    \multirow{2}[2]{*}{Ours(VGG19)}& ZSL   & 30.3   & 33.1  & 33.1 \\
                              & GZSL  & 10.2   & 16.3   & 19.9 \\
    \cmidrule{2-5}
    \multirow{2}[2]{*}{(ML)$^2$P-Encoder(VGG19)}& ZSL   & 29.4   & 32.8  & 32.3 \\
                              & GZSL  & 10.2   & 16.8   & 19.2 \\
    \bottomrule
    \end{tabular}}%
    \vspace{-10px}
  \label{tab1}%
\end{table}

\subsection{Comparison on Different Backbones}
In this section, we compare our method with different backbone networks on the NUS-Wide~\cite{chua2009nus} dataset. We choose the VGG19~\cite{simonyan2014very}, pre-trained on \textit{ImageNet} dataset as the backbone to compare with the best method (ML)$^2$P-Encoder~\cite{liu20232}. From the comparison results, we can see that our model can achieve better performance even if it uses the same backbone network as the comparison method. This result fully demonstrates that our method can efficiently learn single-modal knowledge and also verifies the effectiveness of the design method.

\subsection{Qualitative Results}
\begin{figure*}[htbp]
    \centering
    \includegraphics[width=.98\textwidth]{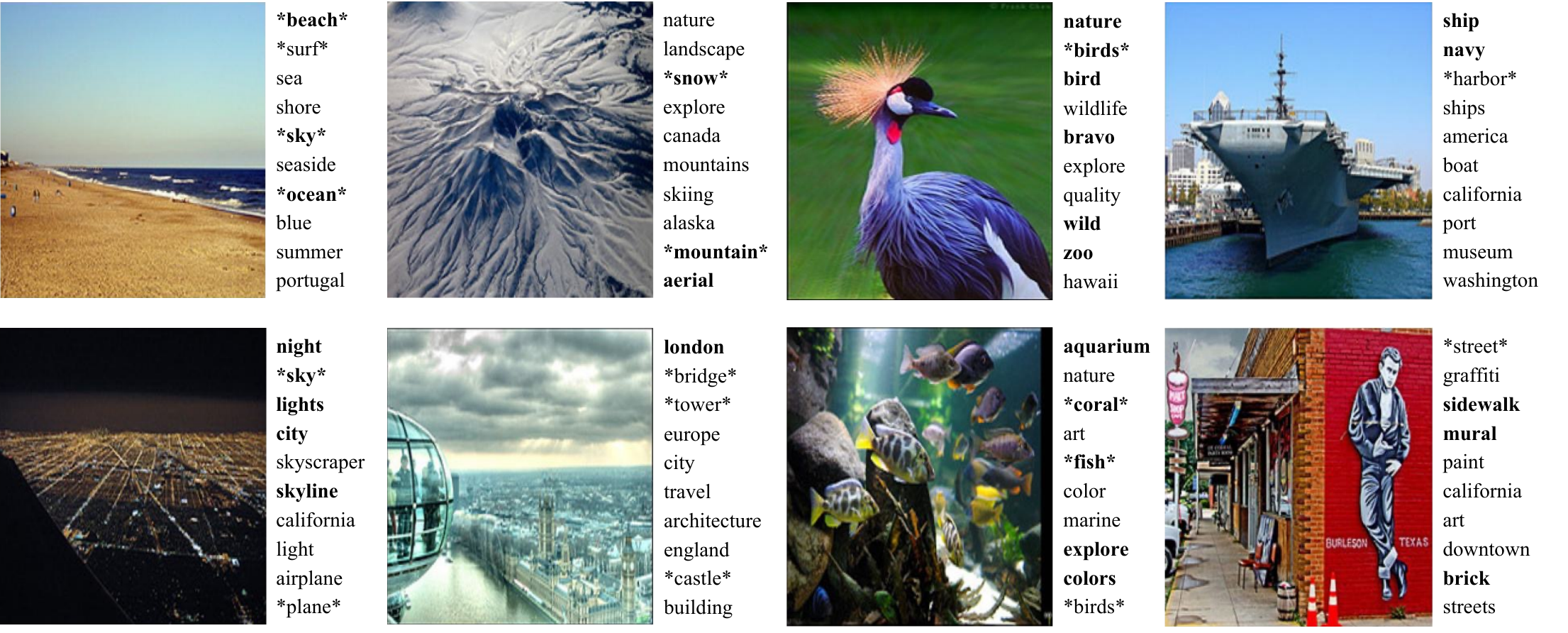}
    \caption{\textbf{Qualitative results.} The top-10 labels predicted by \textit{Epsilon} in the case of Generalized MLZSL on NUS-Wide dataset are shown above. The asterisk marks indicate unseen labels, while the bold ones indicate successfully predicted seen and unseen labels.}
    \label{fig:quali_app}
\end{figure*}

The eight pictures in Figure~\ref{fig:quali_app} respectively show the predictive capabilities of our model under different scenarios. The first and the second pictures (`beach' and `snow') express more of the scene information of the image, so there are extremely high requirements for the model's ability to use global features to make judgments. It can be seen from the prediction results that our model can identify the scene information contained in the image in the first three predicted labels well, which shows that our model is effective in extracting scene information. In the third and fourth pictures, we tend to understand and describe the details of the object. We can see that the third picture is quite accurate for bird identification and prediction. In the fourth picture, we determined through the details of the ship that it belonged to a naval vessel and was anchored in the port. The above-predicted classes all belong to the high response of the model top-3 or even top-5, which is enough to prove that the model accurately mines detailed semantics. The prediction results in the second-row show that our model can still ensure accurate prediction and recognition capabilities (or GZSL capabilities) in the face of more diverse scenes. Including but not limited to the first picture in the second row, the model accurately identified that this is a city at night. But what's especially important to note is that the model even realizes that this is an aerial image. The model in the second picture shows that it is in the UK through the building details in the lower right corner. This ability to master detailed information is amazing. The third picture's grasp of the aquarium scene and the fourth picture's identification of the `red brick wall' both prove the effectiveness of our method from global and local perspectives.

\subsection{Attention Visualization Results}
\begin{figure*}[htbp]
    \centering
    \includegraphics[width=.98\textwidth]{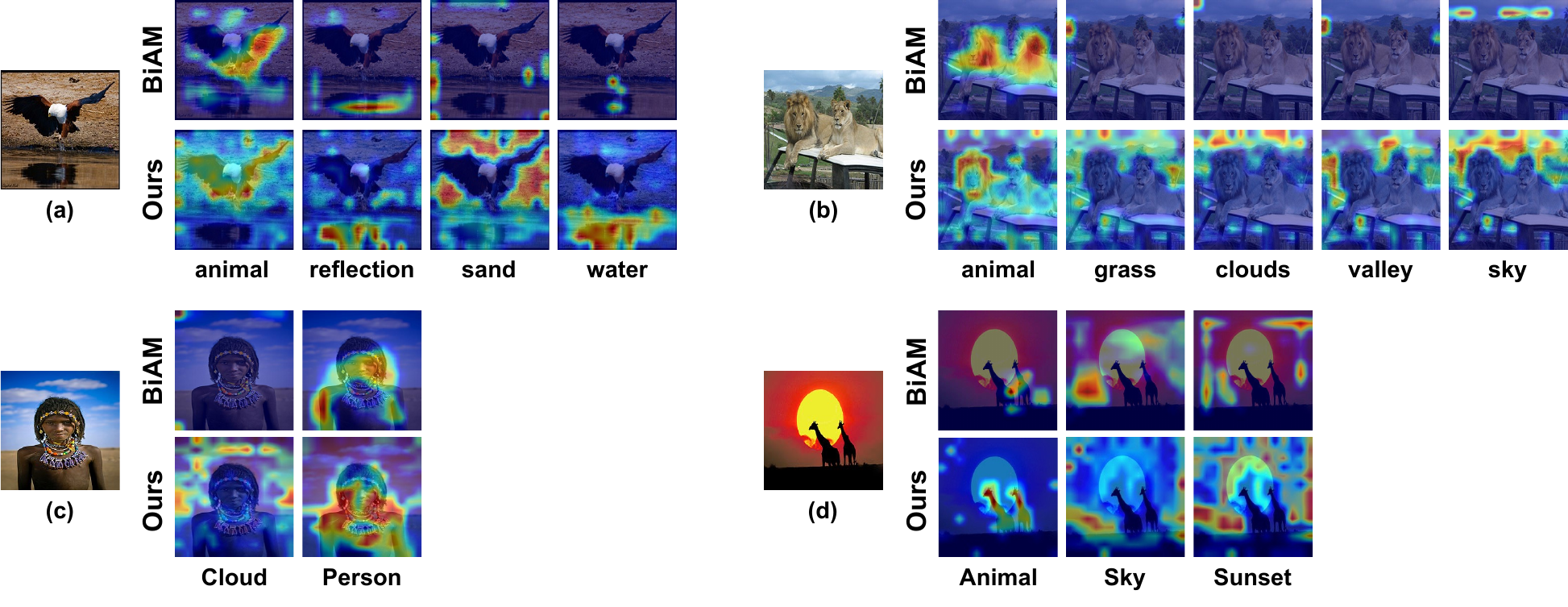}
    \caption{\textbf{Attention visualization of unseen labels in NUS-Wide test-set.} }
    \label{fig:att_app}
\end{figure*}

Figure~\ref{fig:att_app} shows the visual attention results. We still compare with the BiAM~\cite{narayan2021discriminative}. This is because BiAM also claims that its bi-attention approach can integrate global and local features well. However, we can see from the comparison results that BiAM is indeed more accurate in grasping the location of object classes, such as `animal' in (a) and (b). But for scene information, its positioning ability is extremely poor, such as `grass', `cloud', `water', etc. This shows that the model does not take into account global scene information well. On the other hand, our method can grasp the location of these scene classes well and perform accurate positioning. For object classes, the focus of the model is basically accurate. Therefore, our model is able to achieve better results. From Figure~\ref{fig:att_app} (c), we can see that the spatial attention model represented by BiAM has lost almost all scene information, such as the `cloud' of the background. The same problem also occurs in Figure~\ref{fig:att_app} (d). `sky' and `sunset' are both scene information, but our model can perceive all unseen scene classes. For object classes, such as `person' and `animal', our positioning is also more accurate and comprehensive, capturing almost all objects. It can be seen that compared to existing methods, our model can capture both global and local information very robustly.

\bibliography{Epsilon}

\begin{thebibliography}{34}
\providecommand{\natexlab}[1]{#1}

\bibitem[{Akata et~al.(2016{\natexlab{a}})Akata, Malinowski, Fritz, and Schiele}]{akata2016multi}
Akata, Z.; Malinowski, M.; Fritz, M.; and Schiele, B. 2016{\natexlab{a}}.
\newblock Multi-cue zero-shot learning with strong supervision.
\newblock In \emph{Proceedings of the IEEE Conference on Computer Vision and Pattern Recognition}, 59--68.

\bibitem[{Akata et~al.(2016{\natexlab{b}})Akata, Perronnin, Harchaoui, and Schmid}]{akata2016label}
Akata, Z.; Perronnin, F.; Harchaoui, Z.; and Schmid, C. 2016{\natexlab{b}}.
\newblock Label-embedding for image classification.
\newblock \emph{IEEE transactions on pattern analysis and machine intelligence}, 38(7): 1425--1438.

\bibitem[{Ben-Cohen et~al.(2021)Ben-Cohen, Zamir, Ben-Baruch, Friedman, and Zelnik-Manor}]{ben2021semantic}
Ben-Cohen, A.; Zamir, N.; Ben-Baruch, E.; Friedman, I.; and Zelnik-Manor, L. 2021.
\newblock Semantic diversity learning for zero-shot multi-label classification.
\newblock In \emph{Proceedings of the IEEE/CVF International Conference on Computer Vision}, 640--650.

\bibitem[{Chen et~al.(2019)Chen, Wei, Wang, and Guo}]{chen2019multi}
Chen, Z.-M.; Wei, X.-S.; Wang, P.; and Guo, Y. 2019.
\newblock Multi-label image recognition with graph convolutional networks.
\newblock In \emph{Proceedings of the IEEE/CVF Conference on Computer Vision and Pattern Recognition}, 5177--5186.

\bibitem[{Cheng et~al.(2021)Cheng, Lin, Wu, Yang, Shen, Wang, Shi, and Liu}]{cheng2021mltr}
Cheng, X.; Lin, H.; Wu, X.; Yang, F.; Shen, D.; Wang, Z.; Shi, N.; and Liu, H. 2021.
\newblock MlTr: Multi-label Classification with Transformer.
\newblock \emph{arXiv preprint arXiv:2106.06195}.

\bibitem[{Chua et~al.(2009)Chua, Tang, Hong, Li, Luo, and Zheng}]{chua2009nus}
Chua, T.-S.; Tang, J.; Hong, R.; Li, H.; Luo, Z.; and Zheng, Y. 2009.
\newblock Nus-wide: a real-world web image database from national university of singapore.
\newblock In \emph{Proceedings of the ACM international conference on image and video retrieval}, 1--9.

\bibitem[{Deutsch et~al.(2017)Deutsch, Kolouri, Kim, Owechko, and Soatto}]{deutsch2017zero}
Deutsch, S.; Kolouri, S.; Kim, K.; Owechko, Y.; and Soatto, S. 2017.
\newblock Zero shot learning via multi-scale manifold regularization.
\newblock In \emph{Proceedings of the IEEE conference on computer vision and pattern recognition}, 7112--7119.

\bibitem[{Dosovitskiy et~al.(2020)Dosovitskiy, Beyer, Kolesnikov, Weissenborn, Zhai, Unterthiner, Dehghani, Minderer, Heigold, Gelly et~al.}]{dosovitskiy2020image}
Dosovitskiy, A.; Beyer, L.; Kolesnikov, A.; Weissenborn, D.; Zhai, X.; Unterthiner, T.; Dehghani, M.; Minderer, M.; Heigold, G.; Gelly, S.; et~al. 2020.
\newblock An Image is Worth 16x16 Words: Transformers for Image Recognition at Scale.
\newblock In \emph{International Conference on Learning Representations}.

\bibitem[{Durand, Mehrasa, and Mori(2019)}]{durand2019learning}
Durand, T.; Mehrasa, N.; and Mori, G. 2019.
\newblock Learning a deep convnet for multi-label classification with partial labels.
\newblock In \emph{Proceedings of the IEEE/CVF Conference on Computer Vision and Pattern Recognition}, 647--657.

\bibitem[{Feng, An, and He(2019)}]{feng2019collaboration}
Feng, L.; An, B.; and He, S. 2019.
\newblock Collaboration based multi-label learning.
\newblock In \emph{Proceedings of the AAAI Conference on Artificial Intelligence}, volume~33, 3550--3557.

\bibitem[{Frome et~al.(2013)Frome, Corrado, Shlens, Bengio, Dean, Ranzato, and Mikolov}]{frome2013devise}
Frome, A.; Corrado, G.~S.; Shlens, J.; Bengio, S.; Dean, J.; Ranzato, M.; and Mikolov, T. 2013.
\newblock Devise: A deep visual-semantic embedding model.
\newblock \emph{Advances in neural information processing systems}, 26.

\bibitem[{Gao and Zhou(2021)}]{gao2021learning}
Gao, B.-B.; and Zhou, H.-Y. 2021.
\newblock Learning to discover multi-class attentional regions for multi-label image recognition.
\newblock \emph{IEEE Transactions on Image Processing}, 30: 5920--5932.

\bibitem[{Gong et~al.(2013)Gong, Jia, Leung, Toshev, and Ioffe}]{gong2013deep}
Gong, Y.; Jia, Y.; Leung, T.; Toshev, A.; and Ioffe, S. 2013.
\newblock Deep convolutional ranking for multilabel image annotation.
\newblock \emph{arXiv preprint arXiv:1312.4894}.

\bibitem[{Guo and Guo(2020)}]{guo2020novel}
Guo, J.; and Guo, S. 2020.
\newblock A novel perspective to zero-shot learning: Towards an alignment of manifold structures via semantic feature expansion.
\newblock \emph{IEEE Transactions on Multimedia}, 23: 524--537.

\bibitem[{Gupta et~al.(2023)Gupta, Narayan, Khan, Khan, Shao, and Van De~Weijer}]{gupta2023generative}
Gupta, A.; Narayan, S.; Khan, S.; Khan, F.~S.; Shao, L.; and Van De~Weijer, J. 2023.
\newblock Generative multi-label zero-shot learning.
\newblock \emph{IEEE Transactions on Pattern Analysis and Machine Intelligence}.

\bibitem[{Huynh and Elhamifar(2020)}]{huynh2020shared}
Huynh, D.; and Elhamifar, E. 2020.
\newblock A shared multi-attention framework for multi-label zero-shot learning.
\newblock In \emph{Proceedings of the IEEE/CVF Conference on Computer Vision and Pattern Recognition}, 8776--8786.

\bibitem[{Kim, Jun, and Zhang(2018)}]{kim2018bilinear}
Kim, J.-H.; Jun, J.; and Zhang, B.-T. 2018.
\newblock Bilinear attention networks.
\newblock \emph{arXiv preprint arXiv:1805.07932}.

\bibitem[{Kingma and Ba(2014)}]{kingma2014adam}
Kingma, D.~P.; and Ba, J. 2014.
\newblock Adam: A method for stochastic optimization.
\newblock \emph{arXiv preprint arXiv:1412.6980}.

\bibitem[{Kodirov, Xiang, and Gong(2017)}]{kodirov2017semantic}
Kodirov, E.; Xiang, T.; and Gong, S. 2017.
\newblock Semantic autoencoder for zero-shot learning.
\newblock In \emph{Proceedings of the IEEE conference on computer vision and pattern recognition}, 3174--3183.

\bibitem[{Lee et~al.(2018)Lee, Fang, Yeh, and Wang}]{lee2018multi}
Lee, C.-W.; Fang, W.; Yeh, C.-K.; and Wang, Y.-C.~F. 2018.
\newblock Multi-label zero-shot learning with structured knowledge graphs.
\newblock In \emph{Proceedings of the IEEE conference on computer vision and pattern recognition}, 1576--1585.

\bibitem[{Liu et~al.(2023)Liu, Guo, Lu, Guo, Zhang, Zeng, and Huo}]{liu20232}
Liu, Z.; Guo, S.; Lu, X.; Guo, J.; Zhang, J.; Zeng, Y.; and Huo, F. 2023.
\newblock ML2P-Encoder: On Exploration of Channel-Class Correlation for Multi-Label Zero-Shot Learning.
\newblock In \emph{Proceedings of the IEEE/CVF Conference on Computer Vision and Pattern Recognition}, 23859--23868.

\bibitem[{Mensink, Gavves, and Snoek(2014)}]{mensink2014costa}
Mensink, T.; Gavves, E.; and Snoek, C.~G. 2014.
\newblock Costa: Co-occurrence statistics for zero-shot classification.
\newblock In \emph{Proceedings of the IEEE conference on computer vision and pattern recognition}, 2441--2448.

\bibitem[{Narayan et~al.(2021)Narayan, Gupta, Khan, Khan, Shao, and Shah}]{narayan2021discriminative}
Narayan, S.; Gupta, A.; Khan, S.; Khan, F.~S.; Shao, L.; and Shah, M. 2021.
\newblock Discriminative region-based multi-label zero-shot learning.
\newblock In \emph{Proceedings of the IEEE/CVF International Conference on Computer Vision}, 8731--8740.

\bibitem[{Norouzi et~al.(2014)Norouzi, Mikolov, Bengio, Singer, Shlens, Frome, Corrado, and Dean}]{norouzi2014zero}
Norouzi, M.; Mikolov, T.; Bengio, S.; Singer, Y.; Shlens, J.; Frome, A.; Corrado, G.~S.; and Dean, J. 2014.
\newblock Zero-shot learning by convex combination of semantic embeddings.
\newblock In \emph{2nd International Conference on Learning Representations, ICLR 2014}.

\bibitem[{Radford et~al.(2021)Radford, Kim, Hallacy, Ramesh, Goh, Agarwal, Sastry, Askell, Mishkin, Clark et~al.}]{radford2021learning}
Radford, A.; Kim, J.~W.; Hallacy, C.; Ramesh, A.; Goh, G.; Agarwal, S.; Sastry, G.; Askell, A.; Mishkin, P.; Clark, J.; et~al. 2021.
\newblock Learning transferable visual models from natural language supervision.
\newblock In \emph{International conference on machine learning}, 8748--8763. PMLR.

\bibitem[{Simonyan and Zisserman(2014)}]{simonyan2014very}
Simonyan, K.; and Zisserman, A. 2014.
\newblock Very deep convolutional networks for large-scale image recognition.
\newblock \emph{arXiv preprint arXiv:1409.1556}.

\bibitem[{Szegedy et~al.(2015)Szegedy, Liu, Jia, Sermanet, Reed, Anguelov, Erhan, Vanhoucke, Rabinovich et~al.}]{szegedy2015going}
Szegedy, C.; Liu, W.; Jia, Y.; Sermanet, P.; Reed, S.; Anguelov, D.; Erhan, D.; Vanhoucke, V.; Rabinovich, A.; et~al. 2015.
\newblock Going deeper with convolutions.
\newblock Cvpr.

\bibitem[{Tsoumakas and Katakis(2007)}]{tsoumakas2007multi}
Tsoumakas, G.; and Katakis, I. 2007.
\newblock Multi-label classification: An overview.
\newblock \emph{International Journal of Data Warehousing and Mining (IJDWM)}, 3(3): 1--13.

\bibitem[{Veit et~al.(2017)Veit, Alldrin, Chechik, Krasin, Gupta, and Belongie}]{veit2017learning}
Veit, A.; Alldrin, N.; Chechik, G.; Krasin, I.; Gupta, A.; and Belongie, S. 2017.
\newblock Learning from noisy large-scale datasets with minimal supervision.
\newblock In \emph{Proceedings of the IEEE conference on computer vision and pattern recognition}, 839--847.

\bibitem[{Wang et~al.(2016)Wang, Yang, Mao, Huang, Huang, and Xu}]{wang2016cnn}
Wang, J.; Yang, Y.; Mao, J.; Huang, Z.; Huang, C.; and Xu, W. 2016.
\newblock Cnn-rnn: A unified framework for multi-label image classification.
\newblock In \emph{Proceedings of the IEEE conference on computer vision and pattern recognition}, 2285--2294.

\bibitem[{Weston, Bengio, and Usunier(2011)}]{weston2011wsabie}
Weston, J.; Bengio, S.; and Usunier, N. 2011.
\newblock Wsabie: Scaling up to large vocabulary image annotation.
\newblock In \emph{Twenty-Second International Joint Conference on Artificial Intelligence}.

\bibitem[{Xian, Schiele, and Akata(2017)}]{xian2017zero}
Xian, Y.; Schiele, B.; and Akata, Z. 2017.
\newblock Zero-shot learning-the good, the bad and the ugly.
\newblock In \emph{Proceedings of the IEEE Conference on Computer Vision and Pattern Recognition}, 4582--4591.

\bibitem[{Yu et~al.(2014)Yu, Jain, Kar, and Dhillon}]{yu2014large}
Yu, H.-F.; Jain, P.; Kar, P.; and Dhillon, I. 2014.
\newblock Large-scale multi-label learning with missing labels.
\newblock In \emph{International conference on machine learning}, 593--601. PMLR.

\bibitem[{Zhang, Gong, and Shah(2016)}]{zhang2016fast}
Zhang, Y.; Gong, B.; and Shah, M. 2016.
\newblock Fast zero-shot image tagging.
\newblock In \emph{2016 IEEE Conference on Computer Vision and Pattern Recognition (CVPR)}, 5985--5994. IEEE.

\end{thebibliography}
\end{document}